\documentclass[lettersize,journal]{IEEEtran}
\pdfoutput=1
\usepackage{amsmath,amsfonts,amssymb}
\usepackage{algorithmic}
\usepackage{algorithm}
\usepackage{array}
\usepackage[caption=false,font=normalsize,labelfont=sf,textfont=sf]{subfig}
\usepackage{textcomp}
\usepackage{stfloats}
\usepackage{url}
\usepackage{verbatim}
\usepackage{graphicx}
\usepackage{cite}

\newtheorem{definition}{Definition}
\newtheorem{assumption}{Assumption}
\newtheorem{theorem}{Theorem}
\newtheorem{corollary}{Corollary}
\newtheorem{lemma}{Lemma}
\newtheorem{proposition}{Proposition}
\usepackage{booktabs,tabularx}
\newcolumntype{P}[1]{>{\centering\arraybackslash}p{#1}}
\newcolumntype{M}[1]{>{\centering\arraybackslash}m{#1}}
\begin{document}

\title{To transfer or not transfer: Unified transferability metric and analysis}

\author{Qianshan Zhan, Xiao-Jun Zeng*,~\IEEEmembership{Member,~IEEE}
\thanks{*Corresponding author}
\thanks{Qianshan Zhan and Xiao-Jun Zeng are with the Department of Computer Science, University of Manchester, Manchester, United Kingdom (E-mail: qianshan.zhan@postgrad.manchester.ac.uk; x.zeng@manchester.ac.uk).}}



\maketitle

\begin{abstract}
In transfer learning, transferability is one of the most fundamental problems, which aims to evaluate the effectiveness of arbitrary transfer tasks. Existing research focuses on classification tasks and neglects domain or task differences. More importantly, there is a lack of research to determine whether to transfer or not. To address these, we propose a new analytical approach and metric, Wasserstein Distance based Joint Estimation (WDJE), for transferability estimation and determination in a unified setting: classification and regression problems with domain and task differences. The WDJE facilitates decision-making on whether to transfer or not by comparing the target risk with and without transfer. To enable the comparison, we approximate the target transfer risk by proposing a non-symmetric, easy-to-understand and easy-to-calculate target risk bound that is workable even with limited target labels. The proposed bound relates the target risk to source model performance, domain and task differences based on Wasserstein distance. We also extend our bound into unsupervised settings and establish the generalization bound from finite empirical samples. Our experiments in image classification and remaining useful life regression prediction illustrate the effectiveness of the WDJE in determining whether to transfer or not, and the proposed bound in approximating the target transfer risk.
\end{abstract}


\section{Introduction}\label{sec1}
\IEEEPARstart{T}{ransfer} learning has become increasingly popular due to its potential to improve the performance of learning in a data-scare domain (known as the target domain) by leveraging the knowledge from a data-rich domain (known as the source domain) \cite{pan2010survey}. However, the extent of performance improvement could vary widely, even with the possibility of reduced performance. To avoid this, it is important to answer the fundamental question of whether transfer learning should be performed. This is imperative since it can avoid the potential performance loss caused by brute-force transfer, reveal the underlying mechanisms of transfer, and provide insights for designing effective transfer learning models.

The question about whether to transfer drives the investigations of transferability estimation, which aims at developing an analytical approach and metric that is capable of accurately evaluating the effectiveness of arbitrary transfer tasks, without training on the target domain \cite{ibrahim2023newer}. This is important as very often there are few or no label data in target domains and therefore it is challenging and computationally expensive to carry out the empirical valuation in the target domains. To achieve this, various metrics have been proposed. For example, the H-score \cite{bao2019information} estimates the representation transferability by evaluating their discriminatory strength, with the application of source classification model selection for cross-task transfer learning problems. NCE \cite{tran2019transferability} and LEEP \cite{nguyen2020leep} also focus on the cross-task setting in classification. Under the assumption of identical input source and target domains, NCE estimates the transferability using the conditional entropy between discrete target and source labels. LEEP \cite{nguyen2020leep} measures transferability by computing the log-likelihood based on empirical conditional distribution between source and target labels. The calculated LEEP score is used to select the pre-trained classification models. Similar to LEEP, the recently proposed LogME \cite{you2021logme}, PACTran \cite{ding2022pactran} and OTCE \cite{tan2021otce} also focus on pre-trained model selection according to calculated transferability from marginal evidence, the PAC-Bayesian bound of target tasks, and combined cross-domain cross-task differences, respectively.

Despite achieving significant progress in these existing researches, which can be used to select pre-trained source models or understand the relationships between different tasks, existing metrics suffer from limitations. The first limitation is their inability to provide clear guidance on whether to transfer or not for a given target task. One possible reason is that they estimate the transferability by approximating the target performance after transfer learning alone without comparing it to the performance obtained from target data. Such a comparison is critical in determining whether a task is transferable or not, as the performance gain indicates a successful transfer (transferable) and performance loss indicates a failed transfer (non-transferable). Without this comparison, it is difficult to evaluate the effectiveness of a transfer and to provide decision support on whether to transfer or not. As a result, it becomes insufficient to avoid potential performance loss and save computational resources and time.

The second limitation of existing metrics is that they are specialised for downstream classification tasks, which restricts their applicability in regression problems. In fact, transfer learning in regression problems is more difficult, as acknowledged in \cite{jiang2021regressive}. In classification, discrete classification space provides a straightforward way to estimate task transferability by considering the label differences between the target and source class pairs, as existing metrics typically do. However, the regression space is usually continuous and lacks clear pairwise classes, posing a challenge in transferability estimation for regression problems.

The final limitation of existing metrics is their failure to consider domain and task differences simultaneously, which hinders their practical applicability. Domain difference refers to differences in the feature space or feature distributions between domains, while task difference refers to differences in the label space or labelling mechanisms between domains \cite{pan2010survey}. Considering both domain and task differences is necessary since both can occur in transfer learning and deteriorate the transfer performance \cite{pan2010survey}. Unfortunately, existing metrics typically only account for domain difference or task difference but not both. One exception is OTCE \cite{tan2021otce}, which has shown promising results in approximating transfer performance for classification problems by considering both domain and task differences. However, it is unknown whether OTCE is valid in general settings, and thus, there is a need to establish a sound theoretical foundation for this intuitive idea for both regression and classification problems. Based on the above analyses, three gaps are required to be addressed: (1) the lack of research to determine whether to transfer or not; (2) the inability to regression tasks; and (3) the neglect of domain or task differences.

We are motivated to investigate the transferability under a unified setting: classification and regression problems with domain and task differences, and support decision-making on whether to transfer or not. To achieve this, it is not an easy task. Firstly, it is difficult to connect the transferability with domain and task differences simultaneously, particularly with the theoretical guarantees. Secondly, quantifying the task difference is difficult. Existing transferability metrics \cite{tran2019transferability, nguyen2020leep} achieved the quantification with the assumption of the same input source and target domains and thus likely to fail under the domain differences. Theoretical results also quantify the task difference, but they are difficult to be calculated, especially when the target labels are scarce. Thirdly, estimating transferability for regression tasks is even more challenging as continuous regression spaces lack clear pairwise classes, making existing methods that estimate task differences using class pairs unfeasible. As a result, three challenges need to be addressed. The first challenge is to propose a metric to connect transferability with domain difference and task difference based on theoretical guarantees. The second challenge is to estimate the task difference with the consideration of domain difference using limited available target data. The third challenge is to estimate the transferability for both classification and regression tasks.

To this end, we propose a new approach and metric called \textbf{W}asserstein \textbf{D}istance based \textbf{J}oint \textbf{E}stimation (\textbf{WDJE}) as a solution to the challenges of estimating transferability for both classification and regression problems with domain and task differences. By utilizing the Wasserstein distance \cite{flamary2016optimal}, we could measure the divergence between two discrete or continuous distributions, allowing us to estimate domain and task differences for classification and regression problems. The WDJE score provides decision support for transferring or not by comparing the target risk with and without transfer. To enable this comparison without training on the target domain, one key issue is to approximate the target transfer risk. To achieve this, we propose a non-symmetric, easy-to-understand and easy-to-calculate target risk bound that works well even with limited target labels. The proposed bound relates the target risk to the source model performance, Wasserstein distance in target and source feature distributions (domain difference) and Wasserstein distance in target and source labels (task difference). We also extend our target risk bound into unsupervised settings and establish the generalization bound from finite empirical source and target samples.

Our contributions are summarized as follows:
\begin{itemize}
\item[$\bullet$] 
We propose a new analytical transferability approach and metric WDJE that can determine whether to transfer or not for classification and regression problems with domain and task differences. This is the first attempt to estimate transferability for such a unified setting and provide valuable decision support for practitioners.
\item[$\bullet$] 
We compare the target risk with and without transfer to determine whether the task is transferable or not. To enable this comparison, we propose a non-symmetric, easy-to-understand and easy-to-calculate target risk bound that works well with limited or no target labels. We also establish a learning bound from finite empirical source and target samples. These results facilitate the transferability estimation and determination based on theoretical guarantees, and provide theoretical insights into the generalization of target risk.
\item[$\bullet$] 
We illustrate the effectiveness of the proposed WDJE in determining whether to transfer or not by using the proposed consistency index for both classification and regression tasks. Besides, we demonstrate the superior performance of the proposed bound over the state-of-the-art metrics for approximating the target transfer performance in classification and regression problems.
\end{itemize}

The remainder of this paper is organised as follows. Section \ref{sec2} presents related works. Section \ref{sec3} introduces problem formulation and the preliminary of this paper. Section \ref{sec4} presents transferability estimation and analysis. Section \ref{sec5} validates the effectiveness of the proposed WDJE and proposed bound in classification and regression problems. Section \ref{sec6} concludes this paper.

\section{Related work}\label{sec2}
In this section, we provide an overview of three areas in transfer learning that are closely related to our work, i.e., theoretical analysis of transfer learning, empirical transferability estimation and analytical transferability estimation.

\textbf{Theoretical analysis of transfer learning.} Most theoretical studies in transfer learning aim to provide upper bounds for target transfer risks. The main idea is to compare the distribution distances between domains. One of the pioneering theoretical works is proposed by Ben-David et al. \cite{ben2006analysis,ben2010theory}, who introduces the $\mathcal{H} \Delta \mathcal{H}$-divergence as a substitution for traditional distribution divergences such as L1 divergence and KL divergence. This divergence overcomes the difficulties in estimation from finite samples of arbitrary distributions. Using the $\mathcal{H} \Delta \mathcal{H}$-divergence, the generalization error can be upper bounded by the VC-dimension of the hypothesis space. However, this study is limited to binary classification problems and 0-1 loss. To overcome this, Mansour et al. \cite{mansour2009domain} extend it to a general class of loss functions that satisfy the symmetry and subadditivity. They propose the discrepancy distance and establish the data-dependent upper bounds based on Rademacher complexity. To provide theoretical insights for multi-classification problems, Zhang et al. \cite{zhang2019bridging} first introduce disparity discrepancy, which provides a rigorous margin-based generalization bound.

Although target risk bounds can theoretically be used to estimate transferability since they approximate the target performance after a transfer, we empirically find that few studies have achieved this. One possible reason is that these bounds are difficult to compute and abstract, hindering their practical applications. Besides, they are symmetric, which is unsuitable for estimating transferability since transferring from one task to another is different from transferring in the opposite direction. Moreover, these bounds are typically developed for domain adaptation, which is difficult to estimate the task differences. According to these theoretical results, the task difference is modelled through the ideal joint error, a combined error of an ideal joint hypothesis with respect to the source and target labelling mechanisms. However, the ideal joint error cannot be measured directly since it is difficult to obtain the target labelling mechanism and ideal joint hypothesis due to the limited availability of labelled target data.

The analysed limitations suggest that existing target risk bounds are less feasible for transferability estimation and difficult to be applied directly, despite their ability to provide rigorous theoretical guarantees for estimation. Our transferability metric, WDJE, is also derived from the target risk bound. But different from existing bounds, our proposed bound is non-symmetric, easy to understand and compute, and capable of calculating the task difference from limited target labels. For detailed analysis please refer to Section \ref{sec4.2.3}. This makes the proposed bound feasible for practical transferability estimation.

\textbf{Empirical transferability estimation.} Empirical metrics estimate the transferability according to the target performance obtained by re-training or fine-tuning a pre-trained source model. For example, Taskonomy \cite{zamir2018taskonomy} evaluates the transfer performance for each target task by retraining the pre-trained source model to build a non-parametric transferability score. Task2Vec \cite{achille2019task2vec} also requires retraining a probe network on each target task, which takes the task embeddings as input and predicts the target task's outputs. The accuracy of the probe network is then used as the transferability of these embeddings for each task. RSA \cite{dwivedi2019representation} evaluates the transferability of a pre-trained model by comparing the similarity between the representations of the source and target tasks. A higher similarity indicates higher transferability. Generally, these empirical methods often rely on computationally expensive training and may not be practical for practitioners to employ. In contrast, our proposed WDJE provides an analytical approach and metric that does not require finetuning or re-training on the target domain, making it more efficient and more widely applicable in applications.

\textbf{Analytical transferability estimation.} Analytical metrics aim to estimate the transferability efficiently, without training on the target domain. H-score \cite{bao2019information} estimates the representation transferability by assessing their discriminatory strength based on the statics and information theory. It is applied to select the source classification models for cross-task transfer learning problems. Similar to H-score, NCE \cite{tran2019transferability} and LEEP \cite{nguyen2020leep} also focus on the cross-task setting with discrete labels. Under the assumption of identical input domains, NCE estimates the transferability by calculating the conditional entropy between target and source labels, with the application of selecting the highly transferable features from source classification models. LEEP \cite{nguyen2020leep} measures the transferability as the average log-likelihood of the expected empirical predictor, a classifier that makes predictions according to the expected empirical conditional distribution between the source and target labels. The calculated LEEP score is applied to select pre-trained models for classification tasks. The recently proposed PACTran \cite{ding2022pactran}, LogME \cite{you2021logme}, OTCE \cite{tan2021otce}, and GBC \cite{pandy2022transferability} also focus on assessing pre-trained source models. PACTran \cite{nguyen2020leep} estimates transferability from PAC-Bayesian bounds and is applied to select source models for downstream classification tasks. LogME \cite{you2021logme} estimates the transferability by calculating the maximum value of label evidence given the extracted features, making it useful to assess pre-trained models in regression and classification problems. OTCE \cite{tan2021otce} evaluates transferability for classification tasks as a combination of domain difference and task difference based on user-specified parameters. It is applied to select pre-trained models and determine feature weights in multi-source feature fusion for classification tasks. GBC \cite{pandy2022transferability} estimates the transferability by quantifying the amount of overlap between each pair of target classes in the source feature space, where target classes are modelled as Gaussian distribution. It is also applied to select source models and target datasets for classification tasks.

The summarization of these analytical metrics is presented in Table \ref{table1}. Although these metrics achieve impressive performance in various applications such as source model selection and target dataset selection, they suffer from limitations. First, they cannot provide the decision support for whether to perform the transfer or not. Second, most of these metrics are designed for classification problems and are inapplicable for regression problems. Third, these metrics usually account for either domain difference or task difference alone, which are difficult to handle transfer learning scenarios with both domain and task differences. Although OTCE is an exception, which has shown promising results in approximating transfer performance for classification problems. However, it requires parameter tuning and lacks theoretical guarantees for considering both differences simultaneously. Moreover, it is not applicable to regression problems. In contrast, our proposed metric, WDJE is also an analytical metric for transferability estimation and determination that overcomes these limitations. Different from existing metrics, the WDJE could provide the decision support for transferring or not in a unified setting: classification and regression problems with domain and task differences, based on theoretical guarantees.

\begin{table}[htbp]
\centering
\caption{Summarisation of existing analytical transferability metrics.}
\vspace{-2mm}
\small
\renewcommand\arraystretch{1.2}
\setlength{\tabcolsep}{0.4mm}{
\begin{tabular}{lcccc}
\toprule[1pt]         

Metrics    & Account for  & Account for     & Applicable to   & Applicable to  \\
            & domain  & task                &regression      & determine whether \\
           &  differences &  differences   & problems       & to transfer or not\\
\midrule[1pt]
H-score                     & $\times$   & \checkmark & $\times$      & $\times$                      \\
NCE                         & $\times$   & \checkmark & $\times$    & $\times$                  \\
LEEP                        & $\times$  & \checkmark & $\times$   & $\times$                    \\
LogMe                       & \checkmark   & $\times$ & \checkmark      & $\times$                    \\
PACTran                     & $\times$   & \checkmark & $\times$    & $\times$                 \\
OTCE                        & \checkmark  & \checkmark & $\times$  & $\times$                 \\
GBC                         & $\times$ & \checkmark & $\times$   & $\times$ \\
\bottomrule[1pt]
\end{tabular}}
\label{table1}
\end{table}

\section{Problem formulation and preliminary}\label{sec3}
\subsection{Problem formulation}\label{sec3.1}
Let $\mathcal{X} \subseteq \mathbb{R}^d$ be the input or feature space, $\mathcal{Y} \subseteq \mathbb{R}$ be the output or label space, and $f$ be a labelling function mapping $\mathcal{X} \rightarrow \mathcal{Y}$ . $p(x, y)$, $p(x)$, and $p(y)$ represent the joint probability distribution over the feature-label space pair $\mathcal{X} \times \mathcal{Y}$, the marginal probability distribution over $\mathcal{X}$, and the marginal probability distribution over $\mathcal{Y}$, respectively. A domain is denoted as $\mathcal{D} = \left\{\mathcal{X}, p(x)\right\}$. We consider one source domain $\mathcal{D}^S = \left\{\mathcal{X}^S, p^S(x)\right\}$ with marginal probability distribution $p^S(x)$ over $\mathcal{X^S}$, and one target domain $\mathcal{D}^T = \left\{\mathcal{X}^T, p^T(x)\right\}$ with marginal probability distribution $p^T(x)$ over $\mathcal{X^T}$. The joint probability distribution, marginal probability distribution over $\mathcal{Y}$, and labelling function mapping $\mathcal{X}^S \rightarrow \mathcal{Y}^S$ for the source domain $\mathcal{D}^S$ are denoted by $p^S(x, y)$, $p^S(y)$, and $f^S$. Parallel notations $p^T(x, y)$, $p^T(y)$, and $f^T$ are reserved for the target domain $\mathcal{D}^T$.

Suppose that there are $N_S$ labelled source samples $\left\{ \left( \mathbf{x}_i^S,y_i^S \right) \right\}_{i=1}^{N_S}$ , where $y_i^S \; (i = 1, ..., N_S)$ can be continuous or discrete output variables. For target samples, there are two subsets: one with labels and one without labels $\mathbf{H} = \left\{\mathbf{H}_L, \mathbf{H}_U\right\} = \left\{ \left\{ \left( \mathbf{x}_i^T,y_i^T \right) \right\}_{i=1}^{N_{t1}}, \left\{ \mathbf{x}_i^T \right\}_{N_{t1} + 1}^{N_T}\right\}
$, where $y_i^T \; (i = 1, ..., N_{t1})$ can also be a continuous or discrete output variable. The number of samples in $\mathbf{H}_L$ (with labels) and $\mathbf{H}_U$ (without labels) are $N_{t1}$ and $N_T - N_{t1}$ such that $N_{t1} \ll N_{T}$ and $N_{t1} \ll N_{S}$. The associated empirical measures of $p^S(x)$, $p^T(x)$, $p^S(y)$, and $p^T(y)$ are denoted by $\hat{p}^S(x) = \frac{1}{N_S} \sum_{i=1}^{N_S} \vartheta(\mathbf{x}_i^S)$, $\hat{p}^T(x) = \frac{1}{N_T} \sum_{i=1}^{N_T} \vartheta(\mathbf{x}_i^T)$, $\hat{p}^S(y) = \frac{1}{N_S} \sum_{i=1}^{N_S} \vartheta(y_i^S)$, and $\hat{p}^T(y) = \frac{1}{N_{t1}} \sum_{i=1}^{N_{t1}} \vartheta(y_i^T)$, where $\vartheta(\mathbf{x}_i^S)$, $\vartheta(\mathbf{x}_i^T)$, $\vartheta(y_i^S)$, and $\vartheta(y_i^T)$ represents the Dirac function \cite{flamary2016optimal} at locations $\mathbf{x}_i^S$, $\mathbf{x}_i^T$, $y_i^S$, and $y_i^T$, respectively. To arrive at the bounds from limited target samples provided in Theorems \ref{the2} and \ref{the3}, for source target labels $y_i^S \; (i = 1, ..., N_S)$, suppose $(y_1^S, ..., y_{N_{t1}}^S) \sim p^{S1}(y)$  and $(y_{N_{t1}+1}^S, ..., y_{N_S}^S) \sim p^{S2}(y)$ with associated empirical measures $\hat{p}^{S1}(y) = \frac{1}{N_{t1}} \sum_{i=1}^{N_{t1}} \vartheta(y_i^S)$ and $\hat{p}^{S2}(y) = \frac{1}{N_S-N_{t1}} \sum_{i=N_{t1} + 1}^{N_S} \vartheta(y_i^S)$.

A hypothesis is a function $h: \mathcal{X} \rightarrow \mathcal{Y}$ defined in the hypothesis space $\mathcal{H}$. For any distribution $p$ over $\mathcal{X}$, the risk function $\mathcal{R}$ is the expected loss of the hypothesis $h$ on domain $\mathcal{D}$, defined as $\mathcal R_{\mathcal{D}}(h,f) = \mathbb{E}_{x \sim p} \left\{L\left[h(x), f(x)\right]\right\}$ where $L: \mathcal{Y} \times \mathcal{Y} \rightarrow \mathbb{R}$ is a loss function. Suppose that $h(\mathcal{D}^S, \mathcal{D}^T)$ denotes the hypothesis learned with transfer learning based on both target and source data, and $h(\varnothing, \mathcal{D}^T)$ denotes the hypothesis learned without transfer learning based on the target data only. Their corresponding risks on target domain are represented as $\mathcal R_{\mathcal{D}^T}\left[h(\mathcal{D}^S, \mathcal{D}^T),f^T\right]$ and $\mathcal R_{\mathcal{D}^T}\left[h(\varnothing, \mathcal{D}^T),f^T\right]$, respectively. In our transferability estimation, we aim to determine whether to transfer or not by comparing the risk with transfer $\mathcal R_{\mathcal{D}^T}\left[h(\mathcal{D}^S, \mathcal{D}^T),f^T\right]$ and the risk without transfer $\mathcal R_{\mathcal{D}^T}\left[h(\varnothing, \mathcal{D}^T),f^T\right]$. If $\mathcal R_{\mathcal{D}^T}\left[h(\mathcal{D}^S, \mathcal{D}^T),f^T\right] < \mathcal R_{\mathcal{D}^T}\left[h(\varnothing, \mathcal{D}^T),f^T\right]$, the task is transferable as the transfer results in a performance gain. Conversely, if $\mathcal R_{\mathcal{D}^T}\left[h(\mathcal{D}^S, \mathcal{D}^T),f^T\right] > \mathcal R_{\mathcal{D}^T}\left[h(\varnothing, \mathcal{D}^T),f^T\right]$, the task is not transferable as the transfer results in a performance loss.

Notations and corresponding descriptions for this study are presented in Table \ref{table2}.
\begin{table*}[htbp]
\centering 
\caption{Notations in this study}
\vspace{-2mm}
\small
\renewcommand\arraystretch{1.2}
\resizebox{\textwidth}{!}{
\begin{tabular}{ll}
\toprule[1pt]
Notations                 & Descriptions\\
\midrule[1pt]
$\mathcal{D}^S$($\mathcal{D}^T$)                     & Source (Target) domain\\
$\mathcal{X}^S$($\mathcal{X}^T$)           & Source (Target) input or feature space\\
$\mathcal{Y}^S$($\mathcal{Y}^T$)       & Source (Target) output or label space\\
$\left\{ \left( \mathbf{x}_i^S,y_i^S \right) \right\}_{i=1}^{N_S}$     & Source samples\\
$\mathbf{H} = \left\{\mathbf{H}_L, \mathbf{H}_U\right\} = \left\{ \left\{ \left( \mathbf{x}_i^T,y_i^T \right) \right\}_{i=1}^{N_{t1}}, \left\{ \mathbf{x}_i^T \right\}_{N_{t1} + 1}^{N_T}\right\}
$ & Target samples\\
$f^S$ ($f^T$)               & Source (Target) labelling function\\
$h$   & Hypothesis in the hypothesis space $\mathcal{H}$\\
$p^S(x, y)$ ($p^T(x, y)$)                       & Source (Target) joint probability distribution\\
$p^S(x)$ ($p^T(x)$)                        & Source (Target) feature distribution\\
$p^S(y)$ ($p^T(y)$)                  & Source (Target) label distribution\\
$p^{S1}(y)$                        & Source label distribution that $(y_1^S, ..., y_{N_{t1}}^S)$ follow\\
$p^{S2}(y)$                   & Source label distribution that $(y_{N_{t1}+1}^S, ..., y_{N_S}^S)$ follow\\
$\hat{p}^S(x)$ ($\hat{p}^T(x)$)                          & Empirical measure of $p^S(x)$ ($p^T(x)$) \\
$\hat{p}^S(y)$ ($\hat{p}^T(y)$)                          & Empirical measure of $p^S(y)$ ($p^T(y)$) \\
$W\left[p^{S}(y), p^{T}(y)\right]$                          & Wasserstein distance in source and target label distributions $p^{S}(y)$ and $p^{T}(y)$  \\
$W\left[p^{S1}(y), p^{T}(y)\right]$                          & Wasserstein distance in source and target label distributions $p^{S1}(y)$ and $p^{T}(y)$\\
$W\left[p^{S}(x), p^{T}(x)\right]$                         & Wasserstein distance in source and target feature distributions $p^{S}(x)$ and $p^{T}(x)$\\
$W\left[\hat{p}^S(x), \hat{p}^T(x)\right]$                         & Empirical Wasserstein distance in source and target empirical feature distributions $\hat{p}^S(x)$ and $\hat{p}^T(x)$\\
$W\left[\hat{p}^{S1}(y), \hat{p}^{T}(y)\right]$                         & Empirical Wasserstein distance in source and target empirical label distributions $\hat{p}^{S1}(y)$ and $\hat{p}^T(y)$\\
$\mathcal R_{\mathcal{D}^T}\left[h(\varnothing, \mathcal{D}^T),f^T\right]$ & Target risk obtained without transfer \\
$\mathcal R_{\mathcal{D}^T}\left[h(\mathcal{D}^S, \mathcal{D}^T),f^T\right]$ & Target risk obtained with transfer\\
$ \sup \left\{\mathcal R_{\mathcal{D}^T}\left[h(\mathcal{D}^S, \mathcal{D}^T),f^T\right]\right\}$ & Upper bound of the target risk obtained with transfer\\
$Tr_{s,t}$   & WDJE score for a task transferred from $\mathcal{D}_S$ to $\mathcal{D}_T$\\
$\widehat{Tr}_{s,t}$   & Empirical transferability for a task transferred from $\mathcal{D}_S$ to $\mathcal{D}_T$\\
\bottomrule[1pt]
\end{tabular}}
\label{table2}
\end{table*}

\subsection{Wasserstein distance}\label{sec3.2}
One effective way to measure the distance between probability measures is to use the Wasserstein distance, which is defined as the minimum cost of transporting one measure to the other \cite{shen2018wasserstein, santambrogio2015optimal}. The Wasserstein distance can be computed by either solving the Monge problem, which aims to find an optimal transport plan, or solving the Kantorovich problem, which aims to find the optimal coupling of the measures \cite{flamary2016optimal, zhang2019optimal}.

Assume $\Omega_s \subseteq \mathbb{R}^{d_s}$ and $\Omega_t \subseteq \mathbb{R}^{d_t}$ denote two compact metric spaces, and $\mathcal{P}(\Omega_s)$ and $\mathcal{P}(\Omega_t)$ denote the spaces of probability measures over the spaces $\Omega_s$ and $\Omega_t$. Given two probability measures $u \in \mathcal{P}(\Omega_s)$, $v \in \mathcal{P}(\Omega_t)$, the Wasserstein distance computed by solving the Monge problem is presented as
\begin{align}
\label{Wass_monge}
W_p(u, v) 
=& \left[\inf _{T \# u=v} \int_{\Omega_{s}} c(x, T(x)) du(x)\right]^{1 / p} \notag \\
=& \inf _{T \# u=v} \mathbb{E}_{x \sim u}\left[c(x, T(x)\right]^{1 / p},
\end{align}
where $p \ge 1$ and $T$ represents a push-forward or transport map from $u$ to $v$ if $T \# u=v$, and $c: \Omega_s \times \Omega_t \rightarrow \mathbb{R}^{+}$ represents the cost function. $W_p(u, v)$ can be interpreted as the minimum energy required to move a probability mass $u(x)$ from $x$ to $T(x)$.

The Wasserstein distance can also be computed by solving the Kantorovich problem \cite{kantorovich2006translocation}. Let $\Gamma \in \mathcal{P}(\Omega_s \times \Omega_t)$ is a set of all joint probability measures on $\Omega_s \times \Omega_t$ with marginals $u$ and $v$, also called a set of couplings. The Wasserstein distance with order $p$ between $u$ and $v$ can be formalised as:
\begin{align}
\label{Wass_kan}
W_p(u, v) 
=& \left[\inf _{ \pi \in \Gamma(u,v)} \int_{\Omega_{s}\times \Omega_t} c(x_s, x_t) d\pi(x_s, x_t)\right]^{1 / p} \notag \\
=& \inf _{ \pi \in \Gamma(u,v)} \mathbb{E}_{x_s \sim u, x_t \sim v} \left[c(x_s, x_t)\right]^{1 / p}.
\end{align}

\section{Transferability estimation and analysis} \label{sec4}
\subsection{Transferability analysis}\label{sec4.1}
As analysed before, it is crucial in determining whether to transfer or not in transferability estimation. To provide such decision support, we compare the target risk obtained with and without transfer. Such a comparison facilitates the decision-making process, as the performance gain indicates a successful transfer (transferable), while performance loss indicates a failed transfer (non-transferable). The definitions of transferability, a transferable task, and a non-transferable task are presented as follows.
\begin{definition}
\label{def1}
The transferability of a transfer task is defined as a comparison between the target risks obtained with and without transfer. If the target risk of the task obtained with a transfer is smaller than that without a transfer, the task is defined as transferable. If the target risk of the task obtained with a transfer is larger than that without a transfer, the task is defined as non-transferable.
\end{definition}

According to Definition \ref{def1}, if the target risk with transfer $\mathcal R_{\mathcal{D}^T}\left[h(\mathcal{D}^S, \mathcal{D}^T),f^T\right]$ is smaller than that without transfer $\mathcal R_{\mathcal{D}^T}\left[h(\varnothing, \mathcal{D}^T),f^T\right]$, i.e., $\mathcal R_{\mathcal{D}^T}\left[h(\mathcal{D}^S, \mathcal{D}^T),f^T\right] < \mathcal R_{\mathcal{D}^T}\left[h(\varnothing, \mathcal{D}^T),f^T\right]$, the task is transferable as the transfer results in a performance gain. Conversely, if the target risk with transfer $\mathcal R_{\mathcal{D}^T}\left[h(\mathcal{D}^S, \mathcal{D}^T),f^T\right]$ is larger than that without transfer $\mathcal R_{\mathcal{D}^T}\left[h(\varnothing, \mathcal{D}^T),f^T\right]$, i.e., $\mathcal R_{\mathcal{D}^T}\left[h(\mathcal{D}^S, \mathcal{D}^T),f^T\right] > \mathcal R_{\mathcal{D}^T}\left[h(\varnothing, \mathcal{D}^T),f^T\right]$, the task is non-transferable as the transfer results in a performance loss.

Based on the above analysis, we can measure the transferability of a transfer task from $\mathcal{D}_S$ to $\mathcal{D}_T$ as
\begin{align}
\label{Trst_init}
Tr_{s,t}
=& \sup \left\{\mathcal R_{\mathcal{D}^T}\left[h(\mathcal{D}^S, \mathcal{D}^T),f^T\right]\right\} \notag\\
&- \mathcal R_{\mathcal{D}^T}\left[h(\varnothing, \mathcal{D}^T),f^T\right],
 \end{align}
where $\sup \left\{\mathcal R_{\mathcal{D}^T}\left[h(\mathcal{D}^S, \mathcal{D}^T),f^T\right]\right\}$ indicates the upper bound of the target risk achieved after transfer learning. If $Tr_{s,t} <0$, it suggests that the transfer task is worth being conducted because the transfer is likely to generate performance gains. Conversely, if $Tr_{s,t} \ge 0$, it is not recommended to conduct the transfer because the transfer is likely to not provide any benefits.

To calculate $Tr_{s,t}$, one key issue is to measure $\sup \left\{\mathcal R_{\mathcal{D}^T}\left[h(\mathcal{D}^S, \mathcal{D}^T),f^T\right]\right\}$. To achieve this, we perform theoretical analysis to investigate the upper bound of the target risk based on Wasserstein distance for our unified setting: classification and regression problems with domain and task differences, as presented in Section \ref{sec4.2}.
\subsection{Upper bound on the target risk} \label{sec4.2}
This section aims to provide the upper bound of the target risk $\mathcal R_{\mathcal{D}^T}\left[h(\mathcal{D}^S, \mathcal{D}^T),f^T\right]$ to check and help transferability estimation. To achieve this, we introduce the upper bound of target risk based on Wasserstein distance that relates both domain and task differences (Section \ref{sec4.2.1}). Then we establish the generalization bound from an infinite number of empirical source and target samples (Section \ref{sec4.2.2}). Finally, we analyse our proposed target risk bound with the comparison of existing ones (Section \ref{sec4.2.3}).

\subsubsection{Wasserstein distance based target risk bound}
\label{sec4.2.1} 
We adopt the Wasserstein distance to represent the domain difference and task difference owning to its two advantages. First, it can be estimated empirically from the distribution without using non-parametric or semi-parametric methods to smoothen them \cite{flamary2016optimal}. Second, it can provide meaningful distances between the discrete and continuous distributions without the need for shared support \cite{flamary2016optimal}. These advantages allow us to estimate domain and task differences for both classification and regression problems.

To help investigate the upper bound of the target risk, we present an assumption about the loss function and a definition about the probabilistic transfer lipschitzness. 

\begin{assumption} \cite{le2021lamda}
\label{assum1}
Let $L: \mathcal{Y} \times \mathcal{Y} \rightarrow \mathbb{R}$ be the loss function measuring the divergence between two labels. For any $y_1, y_2, y_3 \in \mathcal{Y}$, the loss function is assumed to satisfy the following conditions: \\
(1) triangle inequality, i.e., $L(y_1, y_2) \leq L(y_1, y_3) + L(y_3, y_2)$; \\
(2) symmetric, i.e., $L(y_1, y_2) = L(y_2, y_1)$; \\
(3) bounded, i.e.,  $M := \sup |L(y_1, y_2)| < \infty$; \\
(4) $k$-Lipschitz w.r.t a norm $\left \| \cdot \right \|$, i.e., $|L(y_1, y_2) - L(y_1, y_3)| \leq k\left \| y_2 - y_3 \right \|$.
\end{assumption}

Assumption \ref{assum1} holds throughout the whole study. This assumption is easily satisfied when $L$ is any continuous loss or any bounded loss such as cross-entropy loss or MSE loss.

\begin{definition} \cite{courty2017joint}
\label{def2}
Let $\phi: \mathbb{R} \rightarrow \left[ 0, 1\right]$ be a decreasing function. Suppose that the metric $d$ represents the distance defined on the input space and the metric $L$ represents the difference defined on the output space. A hypothesis $h$ is $\phi$-Lipschitz transferable with respect to the joint distribution $\Gamma \left[p^S(x),p^T(x)\right]$ if for all $\lambda> 0$:
\begin{multline}
\label{phi_lip}
\operatorname{Pr}_{\left(\mathbf{x}^S, \mathbf{x}^T\right) \sim \Gamma \left[p^S(x), p^T(x)\right]} \left\{L\left[h\left(\mathbf{x}^T\right), h\left(\mathbf{x}^S\right)\right] \right. \\
\left. >\lambda d\left(\mathbf{x}^S, \mathbf{x}^T\right) \right\} \leq \phi(\lambda),
\end{multline}
where $\phi(\lambda)$ measures the upper bound on the probability that $L\left[h\left(\mathbf{x}^T\right), h\left(\mathbf{x}^S\right)\right]  >\lambda d\left(\mathbf{x}^S, \mathbf{x}^T\right)$ holds. 
\end{definition}

Based on Assumption \ref{assum1} and Definition \ref{def2}, for an arbitrary pair of a hypothesis and a labelling function $(h, f)$, we can now relate the source and target risk for $(h, f)$ to the Wasserstein distance in source and target feature distributions. The relationship is provided in Lemma \ref{lem1}.
\begin{lemma}
\label{lem1}
Assume that the hypothesis $h$ is $\phi$-Lipschitz transferable. For any $h \in \mathcal{H}$, $\lambda > 0$, and $0< \delta <1$, with a probability of at least $1 – \delta$, the difference between the target and source risks can be bounded by 
\begin{equation}
\left|\mathcal{R}_{\mathcal{D}^T}(h, f)-\mathcal{R}_{\mathcal{D}^S}(h, f)\right| \leq k \lambda W\left[p^S(x), p^T(x)\right]+k M \phi(\lambda),
\end{equation}
\end{lemma}
where $M=\sup \left(\left\|h\left(\mathbf{x}^T\right)-h\left(\mathbf{x}^S\right)\right\|_1\right)$, $k$ is the Lipschitz constant regarding the loss function.

The proof is provided in the supplementary material. In Lemma \ref{lem1}, the term $W\left[p^S(x), p^T(x)\right]$ represents the Wasserstein distance in source and target feature distributions. If the output space $h(\cdot)$ is restricted to $[0, 1]$, $M = 1$. This implies that the difference between $\mathcal{R}_{\mathcal{D}^T}(h, f)$ and $\mathcal{R}_{\mathcal{D}^S}(h, f)$ can be bounded as  
\begin{equation}
\left|\mathcal{R}_{\mathcal{D}^T}(h, f)-\mathcal{R}_{\mathcal{D}^S}(h, f)\right| \leq k \lambda W\left[p^S(x), p^T(x)\right]+k\phi(\lambda),
\end{equation}

For a specific domain $\mathcal{D}^S$ or $\mathcal{D}^T$, the following Lemma \ref{lem2} shows how the risks of different hypotheses and labelling function pairs can be related to the Wasserstein distance in source and target label distributions.
\begin{lemma}
\label{lem2}
For any $h \in \mathcal{H}$, the risk differences of different hypothesis and labelling function pairs $(h, f^S)$ and $(h, f^T)$ in a specific domain $\mathcal{D}$ can be bounded by
\begin{equation}
\left|\mathcal{R}_{\mathcal{D}}(h, f^T)-\mathcal{R}_{\mathcal{D}}(h, f^S)\right| \leq W\left[p^S(y), p^T(y)\right].
\end{equation}
\end{lemma}

The proof is provided in the supplementary material. In Lemma \ref{lem2}, the term $W\left[p^S(y), p^T(y)\right]$ represents the Wasserstein distance in source and target label distributions. 

Using Lemmas \ref{lem1} and \ref{lem2}, we are now ready to establish an upper bound for target risk, as provided in Theorem \ref{the1}.
\begin{theorem}
\label{the1}
Assume that the hypothesis $h$ is $\phi$-Lipschitz transferable. Given the marginal distributions $p^S(x)$, $p^T(x)$, $p^S(y)$, and $p^T(y)$, for any $h \in \mathcal{H}$, $\lambda > 0$, and $0< \delta <1$, with a probability of at least $1 – \delta$, the target expected risk is upper bounded by 
\begin{multline}
\mathcal{R}_{\mathcal{D^T}}\left(h, f^T\right) \leq \mathcal{R}_{\mathcal{D^S}}\left(h, f^S\right)+k \lambda W\left[p^S(x), p^T(x)\right]\\
+W\left[p^S(y), p^T(y)\right]+k M \phi(\lambda).
\end{multline}
\end{theorem}

The proof is provided in the supplementary material. In Theorem \ref{the1}, the item $\mathcal{R}_{\mathcal{D^S}}\left(h, f^S\right)$ denotes the risk of the hypothesis $h$ with respect to the source labelling function $f^S$. The item $W\left[p^S(x), p^T(x)\right]$ represents the Wasserstein distance in source and target feature distributions, quantifying the domain differences. The item $W\left[p^S(y), p^T(y)\right]$ represents the Wasserstein distance in source and target label distributions $p^S(y)$ and $p^T(y)$, quantifying the task differences.

Theorem \ref{the1} provides a bound that relates the target risk with source risk, domain difference, and task difference. While the theorem suggests that the task difference can be estimated from the source and target labels, this estimation can be difficult in transfer learning scenarios with limited or no target labels, which are common in practice. To address this problem, we introduce a theorem that shows how we can make use of the limited target labels to establish an upper bound on the task difference for the target risk.

\begin{theorem}
\label{the2}
    Assume that the hypothesis $h$ is $\phi$-Lipschitz transferable. Given the marginal distributions $p^S(x)$, $p^T(x)$, $p^S(y)$, $p^{S1}(y)$, $p^{S2}(y)$, and $p^T(y)$, for any $h \in \mathcal{H}$, $\lambda > 0$, and $0< \delta <1$, with a probability of at least $1 – \delta$,
    \begin{align}
        W\left[p^S(y), p^T(y)\right] \leq& W\left[p^{S1}(y), p^T(y)\right] \nonumber \\
        &+ \mathbb{E}_{y^S \sim p^{S2}(y)} \left[ \left\| y^S \right\|^p \right]^{1/p},
    \end{align}
    and the target risk is bounded by
    \begin{multline}
        \mathcal{R}_{\mathcal{D^T}}\left(h, f^T\right) \leq \mathcal{R}_{\mathcal{D^S}}\left(h, f^S\right)+k \lambda W\left[p^S(x), p^T(x)\right]\\
        +W\left[p^{S1}(y), p^T(y)\right] \\
        + \mathbb{E}_{y^S \sim p^{S2}(y)} \left[ \left\| y^S \right\|^p \right]^{1/p} 
        + k M \phi(\lambda).
\end{multline}
\end{theorem}
Here, $p \ge 1$ is the parameter used for calculating the Wasserstein distance. $W\left[p^{S1}(y), p^T(y)\right]$ represents the Wasserstein distance in source and target label distributions $p^{S1}(y)$ and $p^T(y)$. $\mathbb{E}_{y^S \sim p^{S2}(y)} \left[ \left\| y^S \right\|^p \right]^{1/p}$ is the expectation of the source label $y^S$ with $y^S \sim p^{S2}(y)$.

The proof is provided in the supplementary material. Theorem \ref{the2} allows us to estimate the task difference from limited target labels. In particular, when no target labels are available in the target domain, Theorem \ref{the2} can be extended to the unsupervised setting, where there is no labelled target data available.

\begin{corollary}
\label{cor1}
Assume that the hypothesis $h$ is $\phi$-Lipschitz transferable. Given the marginal distributions $p^S(x)$, $p^T(x)$, $p^S(y)$, and $p^T(y)$, for any $h \in \mathcal{H}$, $\lambda > 0$, and $0< \delta <1$, with a probability of at least $1 – \delta$,
\begin{align}
        W\left[p^S(y), p^T(y)\right] \leq \mathbb{E}_{y^S \sim p^{S}(y)} \left[ \left\| y^S \right\|^p \right]^{1/p},
    \end{align}
    and the target risk is bounded by
    \begin{multline}
        \mathcal{R}_{\mathcal{D^T}}\left(h, f^T\right) \leq \mathcal{R}_{\mathcal{D^S}}\left(h, f^S\right)+k \lambda W\left[p^S(x), p^T(x)\right]\\
        + \mathbb{E}_{y^S \sim p^{S}(y)} \left[ \left\| y^S \right\|^p \right]^{1/p} 
        + k M \phi(\lambda).
    \end{multline}
\end{corollary}

From Corollary \ref{cor1}, the task difference can be estimated as the expected value of source label distribution, which is often a constant for a given source domain. In this case, our proposed bound only considers the domain difference. This is applicable to transductive transfer learning \cite{pan2010survey} scenarios, where the source and target tasks are the same, but the source and target domains differ.

\subsubsection{Generalization bound from empirical samples} \label{sec4.2.2}
Theorem \ref{the1} gives an upper bound that relates the target risk to the source risk and the Wasserstein distance. However, it does not analyse the generalization bound of the target risk from finite empirical samples. To address this problem, this section establishes the inequality that bounds the target risk by the empirical source risk and empirical Wasserstein distance.

\begin{proposition}
\label{pro1}
Assume that the hypothesis $h$ is $\phi$-Lipschitz transferable. Suppose that $\mathcal{G}^S$ represents a function class defined as $\mathcal{G}^S := \left\{ x \mapsto L\left[h(x),f^S(x)\right]  | h \in \mathcal{H}\right\}$. For any $\lambda > 0$ and $0< \delta <1$, with a probability of at least $1 – \delta$, we have
\begin{multline}
\mathcal{R}_{\mathcal{D}^T}\left(h, f^T\right) 
\leq \mathcal{R}_{\hat{\mathcal{D}}^S}\left(h, f^S\right)+k \lambda W\left[p^S(x), p^T(x)\right] \\
+ W\left[p^{S1}(y), p^T(y)\right]+\mathbb{E}_{y^S \sim p^{S2}(y)} \left[ \left\| y^S \right\|^p \right]^{1/p} \\
 +2 \Re_{N^S}\left(\mathcal{G}^S\right)+M_S \sqrt{\frac{1}{2 N_S} \log \left(\frac{1}{\delta}\right)}+k M \phi(\lambda),
\end{multline}
where $\mathcal{R}_{\hat{\mathcal{D}}^S}\left(h, f^S\right)$ represents the empirical source risk, $\Re_{N^S}\left(\mathcal{G}^S\right)$ represents the Rademacher complexity of the function class $\mathcal{G}^S$ over $N_s$ samples, and $M_s = \sup_{x \in \mathcal{X}, h \in \mathcal{H}} L\left[h(x),f^S(x)\right]$. If the output spaces for a hypothesis $h$ and a label function $f^S$ are restricted into $\left[0, 1\right]$, $M_s = 1$.
\end{proposition}

The proof is provided in the supplementary material. The fifth and sixth terms in proposition \ref{pro1} represent the generalization error on the source domain based on Rademacher complexity \cite{bartlett2002rademacher}. Although Proposition \ref{pro1} provides a bound that relates the target risk to the empirical source risk and source generalization error, the difference between the target risk and the empirical Wasserstein distance is still unknown. To address this, we first present a proposition that demonstrates the empirical concentration result for the Wasserstein distance.

\begin{proposition}
\label{pro2}
Suppose that $p \in (0, d/2)$, $\mu$ represents a probability measure in $\mathbb{R}^d$, and $\hat{\mu} = \frac{1}{N} \sum_{i=1}^N \vartheta(\mathbf{x}_i)$ denotes the associated empirical measure with the $\vartheta(\mathbf{x}_i)$ representing the Dirac function at locations $\mathbf{x}_i \; (i = 1, …, N)$. Assume $M_q(\mu) = \int_{\mathbb{R}^d} \left|x\right|^q d\mu(x) < \infty$ for some $q > p$ and $q \neq d / (d-p)$. Then there exists a constant $\zeta$ depending on $p$, $d$ and $q$ such that for any $0< \delta <1$, with probability at least $1-\delta$, the following holds for all  $h \in \mathcal{H}$:
\begin{align}
W_p(\mu, \hat{\mu}) \leq& \varsigma M_q^{p / q}(\mu)\left(N^{-p / d}+N^{-(q-p) / q}\right) \nonumber \\
&+ B \sqrt{\frac{1}{2 N} \log \frac{1}{\delta}}, 
\end{align}
where $B= \sup _{x_1, x_2 \in \mathcal{X}}\left|h\left(x_1\right)-h\left(x_2\right)\right|$ and $\varsigma M_q^{p / q}(\mu)\left(N^{-p / d}+N^{-(q-p) / q}\right)=\sup \left\{\mathbb{E}\left[W_p(\mu, \hat{\mu})\right]\right\}$.
\end{proposition}

The proof is provided in the supplementary material. Proposition \ref{pro2} demonstrates the convergence of the empirical distribution $\hat{\mu}$ to the associated true distribution $\mu$ in terms of Wasserstein distance, when the sample number increases. Using Propositions \ref{pro1} and \ref{pro2}, we present a theorem that bounds the target risk by the empirical source risk and empirical Wasserstein distance.

\begin{theorem}
\label{the3}
    Assume that the hypothesis $h$ is $\phi$-Lipschitz transferable, $p \in (0, d/2)$, $M_q(\mu) = \int_{\mathbb{R}^d} \left|x\right|^q d\mu(x) < \infty$ for some $q > p$ and $q \neq d / (d-p)$. Given empirical measures of distributions $\hat{p}^S(x)$, $\hat{p}^T(x)$, $\hat{p}^{S1}(y)$, $\hat{p}^{S2}(y)$, and $\hat{p}^T(y)$, for any $h \in \mathcal{H}$, $\lambda > 0$ and $0< \delta <1$, with a probability of at least $1 – \delta$, the target risk is upper bounded by
\begin{multline}
\mathcal{R}_{\mathcal{D}^T}\left(h, f^T\right) \leq \mathcal{R}_{\hat{\mathcal{D}}^S}\left(h, f^S\right)+k \lambda W\left[\hat{p}^S(x), \hat{p}^T(x)\right] \\
+W\left[\hat{p}^{S 1}(y), \hat{p}^T(y)\right]+\mathbb{E}_{y^S \sim \hat{p}^{S 2}(y)}\left[\left\|y^S\right\|^p\right]^{1 / p} \\
+B \sqrt{\frac{1}{2} \log \frac{1}{\delta}}\left(\frac{1}{\sqrt{N_S}}+\frac{1}{\sqrt{N_T}}\right)+B \sqrt{\frac{1}{2 \sqrt{N_{t 1}}} \log \frac{1}{\delta}} \\
+\gamma_X+\gamma_Y +2 \Re_{N^S}\left(\mathcal{G}^S\right) 
+M_S \sqrt{\frac{1}{2 N_S} \log \left(\frac{1}{\delta}\right)}+k M \phi(\lambda),
\end{multline}
where
\begin{align}
\gamma_X =& \sup \left\{\mathbb{E}\left[W\left(p^S(x), \hat{p}^S(x)\right)\right]\right\}  \nonumber \\
&+\sup \left\{\mathbb{E}\left[W\left(p^T(x), \hat{p}^T(x)\right)\right]\right\} \nonumber \\
=&\varsigma M_q^{p / q}\left(p^S(x)\right)\left(N_S^{-p / d}+N_S^{-(q-p) / q}\right) \nonumber \\
&+\varsigma M_q^{p / q}\left(p^T(x)\right)\left(N_T^{-p / d}+N_T^{-(q-p) / q}\right),
\end{align}
and \\
\begin{align}
\gamma_Y=&\sup \left\{\mathbb{E}\left[W\left(p^{S1}(y), \hat{p}^{S1}(y)\right)\right]\right\} \nonumber \\
&+\sup \left\{\mathbb{E}\left[W\left(p^T(y), \hat{p}^T(y)\right)\right]\right\} \nonumber \\
=& \varsigma\left[M_q^{p / q}\left(p^{S1}(y)\right) +M_q^{p / q}\left(p^{T}(y)\right)\right] \left(N_{t 1}^{-p / d} \nonumber \right.\\
&\left. +N_{t 1}^{-(q-p) / q}\right).
\end{align}
\end{theorem}
The proof is provided in the supplementary material. The first, second last and third last items correspond to the empirical source risk accompanied by its generalization error. The remaining terms account for the domain and task differences accompanied by their sampling bounds.

\subsubsection{Analysis of the proposed bound} \label{sec4.2.3}
Our theoretical results emphasise the critical role played by source model performance, domain difference, and task difference. Besides, our proposed bound is more feasible and easier to be implemented for transferability estimation and determination in various transfer learning scenarios.

As analysed in Section \ref{sec2}, previous results are limited by their symmetry, abstraction, difficulty in estimating the task difference from limited labelled target data, and difficulty in the calculation. These limitations make them less feasible and difficult to be applied directly to transferability estimation and determination. Our proposed bound overcomes these limitations as follows: 
\begin{itemize}
\item[$\bullet$] 
It is non-symmetric by incorporating the expectation of source label distribution and the source performance, although still using the symmetric metric of the Wasserstein distance. This is more feasible in transferability estimation since transferring from one task to another is different from transferring in the opposite direction.
\item[$\bullet$] 
It explicitly relates the target risk to both domain and task differences in a unified setting for both classification and regression problems with domain and task differences. This makes the proposed bound easier to understand, and more applicable for various transfer learning tasks.
\item[$\bullet$] 
It provides estimations of task differences from limited labelled target data. This is significant because existing bounds are infeasible to measure the task difference from few or no labelled target data but these are the common application scenarios in transfer learning. The task difference is measured by approximating it as the Wasserstein distance in the source and target label distributions and the expectation of source label distribution. In particular, when no target labels are available, the task difference is approximated by the expectation of source label distribution only. This extension facilitates the transferability estimation and determination for transductive transfer learning scenarios.
\item[$\bullet$] 
It is able to be calculated from limited labelled target data and the calculation is not complex. Wasserstein distance is a key element for bound calculation, which is easy to calculate. We can adopt either the Earth Mover's Distance algorithm \cite{santambrogio2015optimal} or the Sinkhorn algorithm \cite{santambrogio2015optimal} to produce an optimal coupling matrix, followed by calculating the inner product between the coupling matrix and the cost matrix. The calculation process can be significantly streamlined with the use of the POT package \cite{flamary2021pot} in PYTHON.
\end{itemize}

Overall, our proposed bound is a non-symmetric, easy-to-understand and easy-to-calculate target risk bound that works well with limited or no target labels. It provides theoretical insights into target risk generalization, and provides theoretical guarantees for transferability estimation and determination. These characteristics and advantages make it well-suited for transferability estimation and determination in various transfer learning scenarios.

\subsection{Proposed WDJE} \label{sec4.4}
Based on the analysis of transferability and upper bound of $\mathcal R_{\mathcal{D}^T}\left[h(\mathcal{D}^S, \mathcal{D}^T),f^T\right]$, we propose a new metric called \textbf{W}asserstein \textbf{D}istance based \textbf{J}oint \textbf{E}stimation (\textbf{WDJE}) for transferability estimation and determination.
\begin{definition}
\label{def3}
    Assume that the hypothesis $h$ is $\phi$-Lipschitz transferable. Given the marginal distributions $p^S(x)$, $p^T(x)$, $p^{S1}(y)$, $p^{S2}(y)$, and $p^T(y)$, for any $h \in \mathcal{H}$, $\lambda > 0$, and $0< \delta <1$, with a probability of at least $1 – \delta$, the WDJE metric for a task transferred from $\mathcal{D}^S$ to $\mathcal{D}^T$ is defined as 
    \begin{align}
        Tr_{s,t} =& \sup \left\{\mathcal R_{\mathcal{D}^T}\left[h(\mathcal{D}^S, \mathcal{D}^T),f^T\right]\right\} - \mathcal R_{\mathcal{D}^T}\left[h(\varnothing, \mathcal{D}^T),f^T\right] \nonumber \\
 =& \mathcal{R}_{\mathcal{D^S}}\left(h(\varnothing, \mathcal{D}^S), f^S\right)+k \lambda W\left[p^S(x), p^T(x)\right] \nonumber \\
        &+W\left[p^{S1}(y), p^T(y)\right] + \mathbb{E}_{y^S \sim p^{S2}(y)} \left[ \left\| y^S \right\|^p \right]^{1/p} \nonumber \\
        &+ k M \phi(\lambda)- \mathcal R_{\mathcal{D}^T}\left[h(\varnothing, \mathcal{D}^T),f^T\right],
\end{align}
where $p \geq 1$ is the parameter used for calculating the Wasserstein distance.
\end{definition}

In Definition \ref{def3}, $\mathcal{R}_{\mathcal{D^S}}\left(h(\varnothing, \mathcal{D}^S), f^S\right)$ represents the source risk based on source data. $\mathcal R_{\mathcal{D}^T}\left[h(\varnothing, \mathcal{D}^T),f^T\right]$ represents the target risk without transfer, which is obtained using the target data directly. $k \lambda W\left[p^S(x), p^T(x)\right]$ represents the domain difference. $W\left[p^{S1}(y), p^T(y)\right] + \mathbb{E}_{y^S \sim p^{S2}(y)} \left[ \left\| y^S \right\|^p \right]^{1/p}$ represents the task difference, and $k M \phi(\lambda)$ captures the upper bound on the probability that $L\left[h\left(\mathbf{x}^T\right), h\left(\mathbf{x}^S\right)\right]  > \lambda d\left(\mathbf{x}^S, \mathbf{x}^T\right)$ holds for a given source-target sample pairs $\left(\mathbf{x}^S, \mathbf{x}^T\right)$, as stated in Definition \ref{def1}.

The obtained $Tr_{s,t}$ is referred to as the WDJE score, which can be used to determine whether to transfer or not. If the WDJE score is less than 0, i.e., $Tr_{s,t} < 0$, the transfer task is worth performing since the predicted performance gain indicates the transfer is likely to be successful. Conversely, if the WDJE score is greater than 0, i.e., $Tr_{s,t} \ge 0$, the transfer task is not suggested to be conducted because the transfer may not yield any significant benefits.

Overall, our WDJE metric incorporates a more comprehensive estimation across domains and tasks for both classification and regressions, with guaranteed theoretical correctness. It can facilitate the decision-making for whether to perform the transfer or not. Besides, the proposed upper bound can be used to select pre-trained models or domains, similar to existing metrics. These advantages highlight the applicability of the WDJE metric in providing rigorous and insightful decision-making support in practice.

\section{Experiments}\label{sec5}
In this section, we evaluate the performance of the WDJE metric in providing the decision support for performing transfer or not in classification and regression problems. We also validate the performance of the proposed bound on approximating the target risk of transfer tasks by comparing it with the start-of-the-art baselines in both classification and regression problems.
\subsection{Performance evaluation} \label{sec5.1}
\subsubsection{Evaluation of the bound performance} \label{sec5.1.1}
Since our proposed bound can be regarded as an approximation of the target risk, we evaluate its performance by computing the Pearson correlation coefficient \cite{cohen2009pearson} between the calculated bound and the empirical target loss. In this context, the empirical target loss is considered the golden standard for describing the target risk.

To obtain the empirical target loss, one type of method is called re-train head. To achieve this, we fix the feature extractor, which is the portion of the pre-trained source network up to and including the penultimate layer, then train a new head from scratch using the target data, where the head is the last fully connected layer of the target network. Another type of method is called finetune, which updates the feature extractor and head simultaneously.
\subsubsection{Evaluation of the transferability metric} \label{sec5.1.2}
To demonstrate the performance of the proposed WDJE, we propose a consistency index according to the WDJE score $Tr_{s,t}$ and the empirical transferability $\widehat{Tr}_{s,t}$. In this context, the $\widehat{Tr}_{s,t}$ is considered as the golden standard for describing the effectiveness of a transfer obtained after the transfer. If $\widehat{Tr}_{s,t} < 0$, the task is empirically transferable since the empirical evidence shows that the transfer task is beneficial and worth to be conducted. Conversely, if $\widehat{Tr}_{s,t}>0$, the task is empirically non-transferable since the empirical evidence shows that the transfer cannot yield performance improvements after completing the transfer learning process.

Specifically, we calculate the empirical transferability $\widehat{Tr}_{s,t}$ as
\begin{multline}
    \widehat{Tr}_{s,t} = \hat{\mathcal R}_{\mathcal{D}^T}\left[h(\mathcal{D}^S, \mathcal{D}^T),f^T\right]  - \hat{\mathcal R}_{\mathcal{D}^T}\left[h(\varnothing, \mathcal{D}^T),f^T\right],
\end{multline}
where $\hat{\mathcal R}_{\mathcal{D}^T}\left[h(\mathcal{D}^S, \mathcal{D}^T),f^T\right]$ represents the empirical target loss obtained after completing a transfer, and $\hat{\mathcal R}_{\mathcal{D}^T}\left[h(\varnothing, \mathcal{D}^T),f^T\right]$ represents the empirical target loss obtained only from the target data.

Using the calculated WDJE score $Tr_{s,t}$ and the empirical transferability $\widehat{Tr}_{s,t}$, we construct a confusion matrix to facilitate the calculation of the consistency index. The confusion matrix is presented in Table \ref{table3}. 
\begin{table}[htbp]
\centering
\caption{Confusion matrix for transferability estimation.}
\vspace{-2mm}
\small
\renewcommand\arraystretch{1.2}
\setlength{\tabcolsep}{6mm}{
\begin{tabular}{ccc}
\toprule[1pt]
 & $Tr_{s,t} > 0$            & $Tr_{s,t} <0$\\
\midrule[1pt]
$\widehat{Tr}_{s,t}>0$                     & $N_{+,+}$ & $N_{+,-}$                      \\
$\widehat{Tr}_{s,t}<0$                     & $N_{-,+}$ & $N_{-,-}$                  \\       \bottomrule[1pt]
\end{tabular}}
\label{table3}
\end{table}

In Table \ref{table3}, $N_{+,+}$ represents the number of tasks that are empirically non-transferable and predicted as possibly non-transferable. $N_{+,-}$ represents the number of tasks that are empirically non-transferable and predicted as transferable. $N_{-,+}$ represents the number of tasks that are empirically transferable and predicted as possibly non-transferable. $N_{-,-}$ represents the number of tasks that are empirically transferable and predicted as transferable.

Using Table \ref{table3}, the consistency index is defined as follows.
\begin{definition}
\label{def4}
Given the number of tasks that are empirically non-transferable and predicted as transferable $N_{+,-}$, and the number of tasks that are empirically transferable and predicted as transferable $N_{-,-}$, the consistency index $CI$ of a transferability metric is defined as
\begin{align}
    CI = \frac{N_{-,-}}{N_{+,-} + N_{-,-}},
\end{align}
where $N_{+,-} + N_{-,-}$ represents the number of tasks that are predicted as transferable. 
\end{definition}

According to Definition \ref{def4}, the value of $CI$ belongs to $\left[0, 1 \right]$. $CI = 0$ indicates the poor performance of the transferability metric, where none of the predicted transferable tasks is empirically transferable. Conversely, $CI = 1$ indicates the excellent performance of the transferability metric, where all of the predicted transferable tasks are empirically transferable.

\subsection{Domains and tasks setups} \label{sec5.2}
\subsubsection{Domains and tasks for classification} \label{sec5.2.1}
The setting for our validation is presented in Table \ref{table4}, which involves two different source domains (CIFAR 10 and Imagenet-1K) and a single target domain (CIFAR100). We adopt Resnet18 \cite{he2016deep} as source models trained on different source domains. To reduce the cost of feature extraction, we employ a partial-data setting, instead of evaluating the metrics on the full source and target dataset. Specifically, we sample a total of $C*nc$ samples from each dataset, where $C$ denotes the number of classes, and $n_c$ denotes the average number of sampled samples per class. In our experiments, we set $nc$ to 300, 300, and 30 for CIFAR10, CIFAR100, and Imagenet-1K, respectively.

\begin{table}[htbp]
\centering
\caption{Tasks and domains for classification.}
\vspace{-2mm}
\small
\renewcommand\arraystretch{1.2}
\setlength{\tabcolsep}{6mm}{
\begin{tabular}{ccc}
\toprule[1pt]
Tasks & $\mathcal{D}^S$            & $\mathcal{D}^T$\\
\midrule[1pt]
T1                    & CIFAR10 & CIFAR100                      \\
T2                    & Imagenet-1K & CIFAR100                  \\       
\bottomrule[1pt]
\end{tabular}}
\label{table4}
\end{table}

To evaluate the performance of WDJE across a wider variety of downstream tasks, we construct several sub-tasks on CIFAR100. This is achieved by varying the number of considered classes $c$ for different source and target task pairs. Specifically, we draw a subset with top $c$ classes from the pool of 100 classes. Besides, we construct various target tasks by ranging different sampling ratios $r$ to evaluate the metrics on a small data regime. This means we sample the training samples with a ratio of $r$ from the constructed CIFAR100 dataset with $C*n_c$ samples.

Based on the above analysis, we construct two different settings of sub-tasks for each task. The first setting involves 49 subtasks constructed by ranging $r$ from 0.02 to 0.5, with step size 0.01 for considered classes $c = \left\{10, 20, 40\right\}$. The second setting involves 80 subtasks constructed by ranging $c$ from 21 to 100 with the sampling ratio $r = \left\{0.04, 0.08, 0.15\right\}$. By adopting these sub-task settings, we can achieve a more comprehensive and reliable evaluation of the transferability metric across a diverse range of downstream tasks.
\subsubsection{Domains and tasks for regression} \label{sec5.2.2}
For the regression task, we evaluate the metrics using the C-MAPSS dataset \cite{ramasso2014performance}, which is commonly used for the remaining useful life (RUL) prediction of aircraft engines \cite{li2018remaining}. As shown in Table \ref{table5}, the total dataset comprises four subsets with a varying number of working conditions and fault modes. We select two typical subsets for source task: FD001, which contains one working condition and one fault mode, and FD004, which contains six working conditions and two fault modes. We train Attention-based LSTM \cite{chen2020machine} on these subsets and then transfer the pre-trained models to the remaining ones, which differ in their working conditions and fault modes. A detailed description of transfer tasks is presented in Table \ref{table6}.
\begin{table}[htbp]
\centering
\caption{Description of the C-MAPSS dataset.}
\vspace{-2mm}
\small
\renewcommand\arraystretch{1.2}
\setlength{\tabcolsep}{0.7mm}{
\begin{tabular}{ccccc}
\toprule[1pt]         
Data  & No. of training  & No. of testing  & No. of working & No. of fault \\
      & engines          & engines         & conditions     & modes \\
\midrule[1pt]
FD001 & 100                     & 100                     & 1              & 1           \\
FD002 & 260                     & 259                     & 6              & 1           \\
FD003 & 100                     & 100                     & 1              & 2           \\
FD004 & 249                     & 248                     & 6              & 2           \\ 
\bottomrule[1pt]
\end{tabular}}
\label{table5}
\end{table}

\begin{table}[htbp]
\centering
\caption{Description of the C-MAPSS dataset.}
\vspace{-2mm}
\small
\renewcommand\arraystretch{1.2}
\begin{tabular}{ccccccc}
\toprule[1pt]
& T12 & T13 & T14 & T41 & T42 & T43 \\
\midrule[1pt]
$\mathcal{D}^S$ & FD001 & FD001 & FD001 & FD004 & FD004 & FD004 \\
$\mathcal{D}^T$ & FD002 & FD003 & FD004 & FD001 & FD002 & FD003 \\
\bottomrule[1pt]
\end{tabular}
\label{table6}
\end{table}
To evaluate the metric with a small data regime in the regression problem, we construct sub-tasks for each task using different sampling ratios $r$, similar to the classification setup. Specifically, we set the sampling ratio ranging from 0.001 to 0.05 with step size 0.001. This allows us to investigate the performance of transferability metrics comprehensively under a wide range of downstream regression tasks.

\subsection{Validation of the proposed WDJE} \label{sec5.3}
\subsubsection{WDJE validation for classification tasks} \label{sec5.3.1}
In this section, we aim to validate the effectiveness of the proposed metric WDJE in determining whether to perform the transfer or not for transfer tasks T1 and T2 presented in Table \ref{table4}. For each task, we construct 49 subtasks and 80 subtasks as stated in Section \ref{sec5.2.1}. 

To calculate the proposed bound in WDJE, we adopt the cross-entropy loss, whose Lipschitz constant $k$ is calculated as $k=\frac{c-1}{c N_T}\|\mathbf{X}\|$ \cite{yedida2021lipschitzlr}. Here, $c$ is the considered number of classes, $N_T$ is the number of target samples, and $\mathbf{X}$ is the feature matrix of $N_T$ target samples. To obtain the $\lambda$ used in the Probabilistic Lipschitz assumption, we follow the method used in \cite{mansour2009domain} by setting $k \lambda$ to 0.001 and then calculating $\lambda$ as $0.001 / k$. With calculated $\lambda$, we can obtain the value of the function $\phi (\lambda)$ by setting it to $\phi (\lambda) = e^{-\lambda}$. Since the output space is restricted to the interval $\left[0, 1\right]$, we can set the $M$ to 1. To obtain the empirical target loss, we re-train a new classifier by running the SGD for 100 epochs with a learning rate of 0.001 and weight decay of 0.0001. The detailed evaluation results of the proposed WDJE for each task under different settings are presented in Tables \ref{table7}-\ref{table8}.
\begin{table}[htbp]
\centering
\caption{Consistency evaluation of WDJE with ranging $r$ on tasks T1 and T2.}
\vspace{-2mm}
\small
\renewcommand\arraystretch{1.2}
\setlength{\tabcolsep}{4mm}{
\begin{tabular}{c|>{\centering\arraybackslash}m{0.4cm}>{\centering\arraybackslash}m{0.4cm}>{\centering\arraybackslash}m{0.4cm}|>{\centering\arraybackslash}m{0.4cm}>{\centering\arraybackslash}m{0.4cm}>{\centering\arraybackslash}m{0.4cm}}
\hline
Tasks           & \multicolumn{3}{c|}{T1} & \multicolumn{3}{c}{T2} \\ \hline
$c$           & 10 & 20 & 40 & 10 & 20 & 40 \\ 
$N_{+, +}$     & 3 & 0 & 0 & 19 & 0 & 0 \\ 
$N_{+, -}$     & 0 & 0 & 0 & 3 & 1 & 0 \\ 
$N_{-, +}$     & 22 & 0 & 0 & 0 & 0 & 0 \\ 
$N_{-, -}$     & 24 & 49 & 49 & 27 & 48 & 49 \\ 
$CI$          & 0.5217 & 1 & 1 & 1 & 1 & 1 \\ \hline
\end{tabular}}
\label{table7}
\end{table}

\begin{table}[htbp]
\centering
\caption{Consistency evaluation of WDJE with ranging $c$ on tasks T1 and T2.}
\vspace{-2mm}
\small
\renewcommand\arraystretch{1.2}
\setlength{\tabcolsep}{4mm}{
\begin{tabular}{c|>{\centering\arraybackslash}m{0.4cm}>{\centering\arraybackslash}m{0.4cm}>{\centering\arraybackslash}m{0.4cm}|>{\centering\arraybackslash}m{0.4cm}>{\centering\arraybackslash}m{0.4cm}>{\centering\arraybackslash}m{0.4cm}}
\hline
Tasks           & \multicolumn{3}{c|}{T1} & \multicolumn{3}{c}{T2} \\ \hline
$r$           &0.04	&0.08	&0.15	&0.04	&0.08	&0.15 \\ 
$N_{+, +}$     & 0 & 0 & 0 & 0 & 0 & 0 \\ 
$N_{+, -}$     & 0 & 0 & 0 & 0 & 1 & 0 \\ 
$N_{-, +}$     & 0 & 0 & 0 & 0 & 0 & 0 \\ 
$N_{-, -}$     & 80 & 80 & 80 & 80 & 79 & 70 \\ 
$CI$          & 1 & 1 & 1 & 1 & 1 & 1 \\ \hline
\end{tabular}}
\label{table8}
\end{table}
According to the results presented in Tables \ref{table7}-\ref{table8}, we conclude that the consistency index $CI$ is 1 in most situations. This implies that the WDJE score achieves high consistency with the empirical transferability, that is, predictions regarding transferable transfer tasks are consistently in line with the empirical validations of transferable transfer tasks. This demonstrates the effectiveness of the WDJE as a transferability estimation metric, which facilitates the decision-making about whether to perform the transfer or not.

\begin{figure}[ht]
\centering
\includegraphics[width=\linewidth]{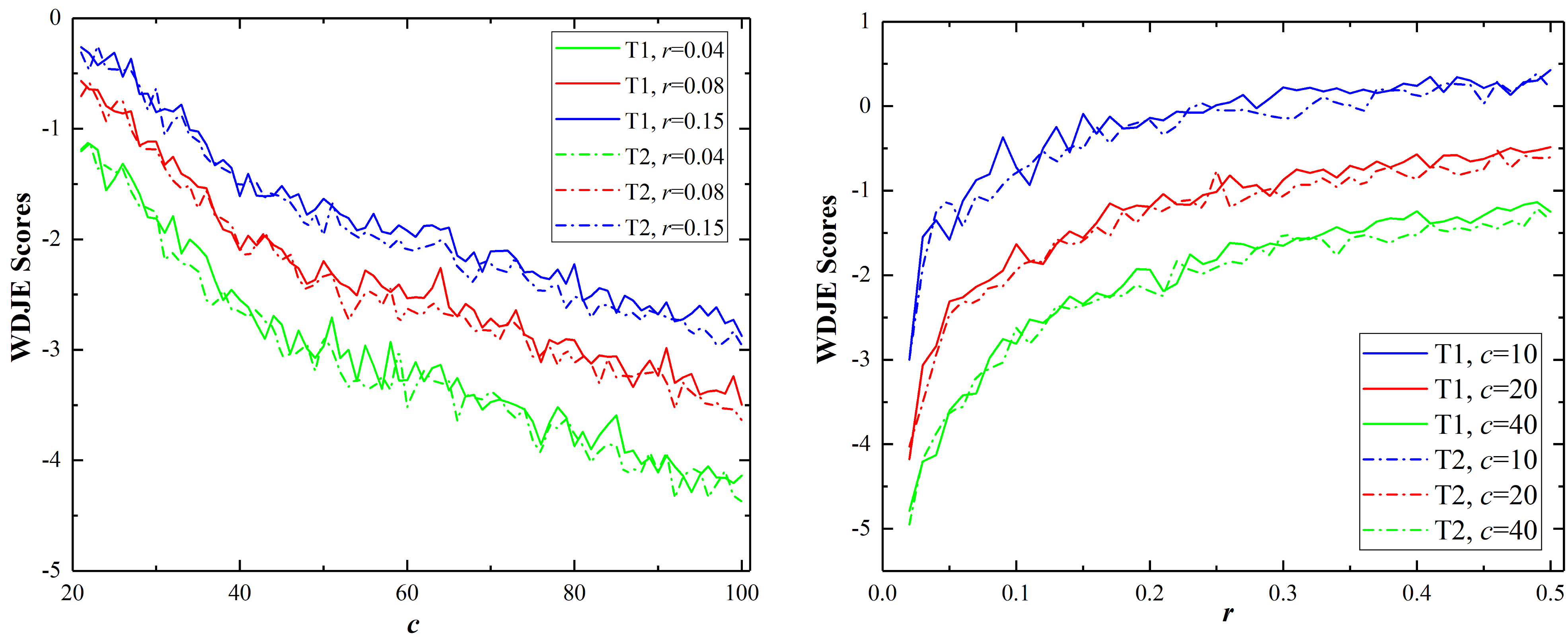}
\caption{Changes in WDJE scores with ranging $c$ and $r$.}
\label{fig1}
\centering
\end{figure}

To illustrate the impact of $c$ and $r$ on the WDJE scores for various tasks, we visualize the changes in WDJE scores across different settings, as depicted in Fig. \ref{fig1}.

From Fig. \ref{fig1}, we found that when the sampling ratio $r$ is fixed, the transferability is increasing with the increase of the considered classes $c$. When $c$ is fixed, the transferability is increasing with the decrease of $r$. This is because when $c$ increases, the classification task becomes more complex to learn, making it more beneficial to use transfer learning instead of training a model from scratch using target data. When $r$ decreases, the classification task becomes more difficult with fewer target data, making it more challenging to train a well-performed model using target data only. In this case, transfer learning can be useful for leveraging knowledge from the source domain to improve the performance of the target model.

Overall, the decision of using transfer learning or not depends on specific situations, if the task is complex and challenging, such as having more classes or fewer target data, transfer learning is effective to improve the target performance. However, if the task is easier, such as having fewer classes or more samples, transfer learning may not be necessary since a model trained on the target domain alone may be sufficient to achieve good performance. This highlights our motivation, that is, it is important to analyse the task and data to determine whether to conduct a transfer or not. 

\subsubsection{WDJE validation for regression tasks} \label{sec5.3.2}
In this section, we validate the performance of the WDJE in determining whether to perform the transfer or not for transfer tasks presented in Table \ref{table5}. For each task, we construct 49 subtasks as stated in Section \ref{sec5.2.2}. For calculating the bound in WDJE, we adopt the MSE loss, with the Lipschitz constant $k$ calculated as $k = \frac{K}{N_{t1}}\|\mathbf{X^T X}\| - \frac{1}{N_{t1}}\|\mathbf{y^T X}\|$ \cite{yedida2021lipschitzlr}. Here, $K$ is the supremum of the weights in the regressor, assumed to be 1 as all weights are normalised to the interval [0, 1]. $N_{t1}$ denotes the number of labelled target samples, and $\mathbf{X}$ and $\mathbf{y}$ represent the feature matrix and label vector of $N_{t1}$ target samples. We maintain the remaining parameters for calculating the bound, as stated in Section \ref{sec5.3.1}. To obtain the empirical target loss, we finetune the source-trained Attention-based LSTM using the Adam optimizer with a learning rate of 0.001 or 0.01 for 100 epochs. The detailed evaluation results for each task under different settings are presented in Table \ref{table9}.
\begin{table}[htbp]
\centering
\caption{Consistency evaluation of WDJE in RUL prediction.}
\vspace{-2mm}
\small
\renewcommand\arraystretch{1.2}
\setlength{\tabcolsep}{3.4mm}{
\begin{tabular}{ccccccc}
\hline
Tasks          & T12 & T13 & T14 & T41 & T42 & T43 \\ \hline
$N_{+, +}$     & 0 & 0 & 0 & 48 & 0 & 46 \\ 
$N_{+, -}$     & 9 & 9 & 5 & 0 & 0 & 2 \\ 
$N_{-, +}$     & 0 & 0 & 0 & 1 & 0 & 1 \\ 
$N_{-, -}$     & 40 & 40 & 44& 0 & 49 & 0 \\ 
$CI$          & 1 & 1 & 1 & 0 & 1 & 0 \\ \hline
\end{tabular}}
\label{table9}
\end{table}

According to the results, we conclude that the WDJE score achieves a high level of consistency with empirical transferability. This means that predictions made using the WDJE metric for transferable transfer tasks are consistently in line with the empirical validations of such tasks. For tasks T41 and T43, the $CI$s of the WDJE metric are 0. In these tasks, only one sub-task is found to be empirically transferable among all subtasks. Therefore, the ineffectiveness of WDJE cannot be conclusively deducted from only one empirically transferable task. To sum up, the WDJE can provide effective transferability estimation and determination to support the decision-making regarding whether to transfer or not in regression problems.

Besides, results shown in Table \ref{table9} indicate that WDJE scores exhibit negative values in most scenarios when the source domain is FD001, suggesting good transferability for these tasks. Conversely, when the source domain is FD004, the WDJE scores show to be positive in most scenarios, indicating poor transferability. This may be attributed to the performance of the source model, as the Attention-based LSTM achieves significantly better performance in FD001 ($MSE$ = 14.268) than in FD004 ($MSE$ = 23.353). This highlights the importance of the source model performance, validating the effects of the source model performance in our proposed bound.

Interestingly, when the source domain is FD004, the WDJE scores show to be negative in most scenarios except for the target domain FD002. This may be because there are 6 working conditions shared by FD004 and FD002, resulting in a more similar feature space than that in tasks T41 or T43. This highlights the importance of the domain difference, validating the effects of the domain difference in our proposed bound.

\subsection{Validation of the proposed bound} \label{sec5.4}
\subsubsection{Bound validation for classification tasks} \label{sec5.4.1}
In this section, we validate the effectiveness of the proposed bound in approximating the target risk of transfer tasks T1 and T2. For each task, we construct 49 subtasks and 80 subtasks as stated in Section \ref{sec5.2.1}. To provide a comprehensive evaluation of the proposed bound, we compare it with other metrics for predicting or approximating the target risk, namely the analytical transferability metrics including LEEP \cite{nguyen2020leep}, NCE \cite{tran2019transferability}, LogME \cite{you2021logme}, H-score \cite{bao2019information}, PAC\_gam \cite{ding2022pactran}, and PAC\_dir \cite{ding2022pactran}. We perform this comparison by calculating the Pearson correlation coefficient between the empirical loss and approximations of the target risk derived from the proposed bound and other transferability metrics, as stated in Section \ref{sec5.2.1}. Parameter settings from Section \ref{sec5.3.1} are still used for bound validation. Results about the correlation for each task under different settings are presented in Tables \ref{table10}-\ref{table13}. All the correlation coefficients are statistically significant with $p < 0.001$. The best results are highlighted in bold.
\begin{table}[htbp]
\centering
\caption{Comparisons of correlation coefficients with ranging $c$ on task T1.}
\vspace{-2mm}
\small
\renewcommand\arraystretch{1.2}
\setlength{\tabcolsep}{0.6mm}{
\begin{tabular}{cccccccc}
\hline
$r$                & LEEP   & NCE     & LogME           & H-score & PAC\_gam & PAC\_dir & Proposed\\ \hline
0.04 & 0.9961 & -0.9928 & 0.9983          & 0.9841  & -0.9881  & -0.9877  & \textbf{0.9915} \\
0.08 & 0.9973 & -0.9779 & \textbf{0.9981} & 0.9849  & -0.9867  & -0.9864  & 0.9908          \\
0.15 & 0.9979 & -0.9566 & \textbf{0.998}  & 0.962   & -0.9851  & -0.9846  & 0.991  \\ \hline
\end{tabular}}
\label{table10}
\end{table}

\begin{table}[htbp]
\centering
\caption{Comparisons of correlation coefficients with ranging $r$ on task T1.}
\vspace{-2mm}
\small
\renewcommand\arraystretch{1.2}
\setlength{\tabcolsep}{0.6mm}{
\begin{tabular}{cccccccc}
\hline
$c$                & LEEP   & NCE     & LogME           & H-score & PAC\_gam & PAC\_dir & WDJE\\ \hline
10 & -0.9013 & 0.8274 & 0.7017          & -0.5534 & 0.4611   & 0.7801   & \textbf{0.9499} \\
20 & -0.9542 & 0.8277 & 0.707  & 0.2346  & 0.5033   & 0.634    & \textbf{0.9782} \\
40 & -0.9531 & 0.8154 & 0.7324 & 0.6266  & 0.5615   & 0.6536   & \textbf{0.9871} \\ \hline
\end{tabular}}
\label{table11}
\end{table}

\begin{table}[htbp]
\centering
\caption{Comparisons of correlation coefficients with ranging $c$ on task T2.}
\vspace{-2mm}
\small
\renewcommand\arraystretch{1.2}
\setlength{\tabcolsep}{0.6mm}{
\begin{tabular}{cccccccc}
\hline
$r$                & LEEP   & NCE     & LogME           & H-score & PAC\_gam & PAC\_dir & 
WDJE\\ \hline
0.04 & 0.9868          & -0.8431 & 0.9881 & 0.2106  & -0.9773  & -0.9775  & \textbf{0.9884} \\
0.08 & \textbf{0.9887} & -0.9423 & 0.9882 & 0.8033  & -0.9831  & -0.9831  & 0.9877          \\
0.15 & \textbf{0.9922} & -0.9547 & 0.9917 & 0.8061  & -0.9829  & -0.9829  & 0.9912  \\ \hline
\end{tabular}}
\label{table12}
\end{table}

\begin{table}[htbp]
\centering
\caption{Comparisons of correlation coefficients with ranging $r$ on task T2.}
\vspace{-2mm}
\small
\renewcommand\arraystretch{1.2}
\setlength{\tabcolsep}{0.6mm}{
\begin{tabular}{cccccccc}
\hline
$r$                & LEEP   & NCE     & LogME           & H-score & PAC\_gam & PAC\_dir & WDJE\\ \hline
10 & -0.9285 & 0.6844 & 0.7936 & 0.274   & 0.9459   & 0.9508          & \textbf{0.9768} \\
20 & -0.8589 & 0.7447 & 0.754  & 0.7202  & 0.9594   & 0.9542          & \textbf{0.9618} \\
40 & -0.9174 & 0.7689 & 0.6307 & 0.9065  & 0.9548   & \textbf{0.9577} & 0.9523  \\ \hline
\end{tabular}}
\label{table13}
\end{table}
From Tables \ref{table10}-\ref{table13}, we could find that our proposed bound shows a strong correlation with the empirical target loss after transfer learning, achieving the highest correlation in most cases. Although our proposed bound does not achieve the highest correlation in a few cases, the values are higher than 0.95 and very close to the highest result, differing only in the third decimal place. This is sufficient to be used in practice. All these demonstrate the effectiveness of the proposed bound in accurately approximating the empirical target loss after transfer.
\subsubsection{Bound validation for regression tasks} \label{sec5.4.2}
In this section, we assess the effectiveness of the proposed bound in approximating the target risk for RUL prediction tasks. For each task, we construct 49 subtasks as stated in Section \ref{sec5.2.2}. To further evaluate the performance of the proposed bound, similar to that in classification, we compare it with LogME \cite{you2021logme}, the only transferability metric we have found that is applicable to regression tasks. We perform this comparison by calculating the Pearson correlation coefficient between the empirical loss and approximations of the target risk. Parameter settings from Section \ref{sec5.3.2} are still used for bound validation. Results about the correlation for each task under different settings are presented in Table \ref{table14}. All the correlation coefficients are statistically significant with $p < 0.001$. The best results are highlighted in bold.
\begin{table}[htbp]
\centering
\caption{Comparisons of correlation coefficients in RUL prediction tasks.}
\vspace{-2mm}
\small
\renewcommand\arraystretch{1.2}
\setlength{\tabcolsep}{1.4mm}{
\begin{tabular}{cccccccc}
\hline
Tasks          & T12            & T13             & T14              & T41             & T42            & T43\\ \hline
LogME          & -0.5098         & -0.8195         & \textbf{-0.9032} & -0.9459         & \textbf{-0.485} & -0.9311         \\
WDJE & \textbf{0.5563} & \textbf{0.8452} & 0.8939           & \textbf{0.9727} & 0.4811          & \textbf{0.9612}  \\ \hline
\end{tabular}}
\label{table14}
\end{table}

The results from Table \ref{table14} show that our proposed bound is highly effective in approximating the empirical target loss, with higher correlation coefficients than other metrics in all cases except for tasks T14 and T42. However, even in these cases, the difference is negligible. Overall, our comparative results confirm that the proposed bound can provide an effective approximation of target risks in RUL prediction tasks.

\section{Conclusion} \label{sec6}
Estimating and determining the transferability between the source and target tasks is an essential task and a fundamental challenge in transfer learning to avoid failed transfer and reveal knowledge transfer mechanisms. In this study, we develop an alternative approach and propose a new analytical metric called Wasserstein Distance based Joint Estimation (WDJE) for transferability estimation and determination in a unified setting: classification and regression problems with domain and task differences. The WDJE metric provides the decision support for performing the transfer or not by comparing the target risk obtained with and without transfer. To enable this comparison without training on the target domain, it is essential to approximate the target risk after transfer. To achieve this, we propose a non-symmetric, easy-to-understand and easy-to-calculate target risk bound that works well even with limited target labels. The proposed bound relates the target risk to the source model performance, domain and task differences based on the Wasserstein distance. We also extend our bound into unsupervised settings and establish the generalization bound from finite empirical source and target samples. Our experiments in image classification and RUL regression prediction demonstrate that the WDJE metric provides effective decision support regarding whether to transfer or not and the superior performance of the proposed bound over the state-of-the-art metrics for approximating the target transfer performance. These results highlight the practical value of the WDJE and the proposed bound in classification and regression problems.

In the future, we will extend the proposed metric in multi-source transfer learning scenarios. Besides, we will explore more applications of the WDJE metric, such as leveraging the proposed bound to guide the model design and training process for regression and classification problems with domain and task differences.





\bibliographystyle{IEEEtran}
\bibliography{main}


 




\vfill

\end{document}